\def\year{2018}\relax
\documentclass[letterpaper]{article} 
\usepackage{aaai18}  
\usepackage{times}  
\usepackage{helvet}  
\usepackage{courier}  
\usepackage{url}  
\usepackage{graphicx}  
\usepackage{MyDefinitions}
\begin{document}
%
\title{A spatio-temporalisation of $\alcd$ and its translation into alternating automata
augmented with spatial constraints}
\author{Amar Isli\\
        University of Sciences and Technology Houari Boumedi\`ene\\
	Department of Computer Science\\
	BP 32, DZ-16111 Bab Ezzouar, Algiers\\
	Algeria\\
	a\_isli@yahoo.com\\
}
\maketitle
\begin{abstract}
\footnote{Exactly as rejected by the KR'2018 Conference. The paper, together with another, also rejected by the KR'2018 Conference, had been extracted
from a substantial revision of \cite{Isli03a}. Further revisions are needed before replacing \cite{Isli03a}.}\\
The aim of this work is to provide a family of qualitative theories for spatial change
in general, and for motion of spatial scenes in particular.
To achieve this, we consider a
spatio-temporalisation $\xdl$, of the well-known $\alcd$ family of Description Logics
(DLs) with a concrete domain:
the $\xdl$ concepts are interpreted over infinite $k$-ary $\Sigma$-trees, with the nodes
standing for time points, and $\Sigma$ including, additionally to its uses in classical
$k$-ary $\Sigma$-trees, the description of the snapshot of an $n$-object spatial scene of interest;
the roles split into $m+n$ immediate-successor (accessibility) relations, which are
serial, irreflexive and antisymmetric, and of which $m$ are general, not necessarily functional,
the other $n$ functional;
the concrete domain $\textsl{D}_x$ is generated by an \mbox{$\rcc8$-like} spatial
Relation Algebra (RA) $x$, and is used to guide the change by imposing spatial constraints on objects of
the "followed" spatial scene, eventually at different time points of the input trees.
In order to capture the expressiveness
of most modal temporal logics encountered in the literature, we introduce weakly cyclic Terminological
Boxes (TBoxes) of $\xdl$, whose axioms capture the decreasing property of modal temporal operators.
We show the important result that satisfiability of an $\xdl$ concept with respect to a weakly cyclic TBox
can be reduced to the emptiness problem of a B\"uchi weak alternating automaton augmented with
spatial constraints.
%
%
In another
work, complementary to this one, also submitted to this conference, we thoroughly investigate B\"uchi
automata augmented with spatial constraints, and provide, in particular, a translation of an alternating
into a nondeterministic, and an effective decision procedure for the emptiness problem of the latter.

{\bf Author keywords:}
Spatio-temporal reasoning,
Description logics with a concrete domain,
Weakly cyclic TBox,
Modal temporal logics,
Constraint-based qualitative spatial reasoning,
Alternating automata augmented with constraints.
\end{abstract}
\newtheorem{theorem}{Theorem}
\newtheorem{conjecture}{Conjecture}
\newtheorem{corollary}{Corollary}
\newtheorem{proposition}{Proposition}
\newtheorem{lemma}{Lemma}
\newtheorem{discussion}{Discussion}
\newtheorem{definition}{Definition}
\newtheorem{remark}{Remark}
\newtheorem{example}{Example}
\section{Introduction}\label{sect1}
The goal of the present work is to enhance the expressiveness of modal
temporal logics with qualitative spatial constraints. What we get is a
family of qualitative theories for spatial change in general, and for
motion of spatial scenes in particular. The family consists of
domain-specific spatio-temporal (henceforth $\st$) languages, and is
obtained by spatio-temporalising a well-known family of description logics (DLs)
with a concrete domain, known as $\alcd$ \cite{BaaderH91a}. $\alcd$
originated from a pure DL known as $\alc$
\cite{Schmidt-SchaussS91a}, with $m\geq 0$ roles all of which are
general, not necessarily functional relations, and which Schild
\cite{Schild91a} has shown to be expressively equivalent to Halpern
and Moses' $\textsl{K}_{(m)}$ modal logic \cite{HalpernM85a}. $\alcd$
is obtained by adding to $\alc$
functional roles (better known as abstract features), a concrete
domain $\textsl{D}$, and concrete features (which refer to objects of
the concrete domain). The spatio-temporalisation of $\alcd$ is
obtained, as the name suggests, by performing two specialisations at the same
time:
(1) temporalisation of the roles, so that they consist of $m+n$ immediate-successor (accessibility) relations
    $R_1,\ldots R_m,f _1,\ldots ,f _n$, of which the $R_i$'s are general, the $f_i$'s
    functional; and
(2) spatialisation of the concrete domain $\textsl{D}$: the concrete domain
    is now $\textsl{D}_x$, and is  generated  by a spatial RA $x$, such as the
    Region-Connection Calculus RCC8 \cite{RandellCC92a}.

The final spatio-temporalisation of $\alcd$ will be
referred to as $\xdl$ ($\mtalc$ for \mbox{$\textsl{M}$odal}
\mbox{$\textsl{T}$emporal} $\alc$). 
Constraint-based languages candidate for generating a concrete domain
for a member of our family of spatio-temporal theories, are spatial RAs
for which the atomic relations form a decidable subset ---i.e., such
that consistency of a CSP expressed as a conjunction of $p$-ary
relations on $p$-tuples of objects, where $p$ is the arity of the RA
relations, is decidable. These
include,
the Region-Connection Calculus $\rcc8$ in \cite{RandellCC92a} (see also
\cite{Egenhofer91a}),
the Cardinal Directions Algebra $\cdalg$ in \cite{Frank92a},
and the rectangle algebra in \cite{BalbianiCdC98a} (see also
\cite{Guesgen89a,MukerjeeJ90a}), for the binary case;
and
the RA $\atra$ of 2D orientations in \cite{IsliC98a,IsliC00a} for the
ternary case.
As our illustrating spatial RA, we will be using the ternary RA in \cite{IsliC98a,IsliC00a}.

It is known that, in the general case,
satisfiability of an $\alcd$ concept with respect to a cyclic
Terminological Box (TBox) is undecidable (see, e.g., \cite{Lutz01a}).
In order to capture the expressiveness
of most modal temporal logics encountered in the literature, we introduce in this work weakly cyclic
TBoxes of $\xdl$, whose axioms capture the decreasing property of modal temporal operators.
We show the important result that satisfiability of an $\xdl$ concept with respect to a weakly cyclic TBox
can be reduced to the emptiness problem of a B\"uchi weak alternating automaton augmented with
spatial constraints.
%
%
In another
work, complementary to this one, also submitted to this conference, we thoroughly investigate B\"uchi
automata augmented with spatial constraints, and provide, in particular, a translation of an alternating
into a nondeterministic, and an effective decision procedure for the emptiness problem of the latter.
\section{The $\xdl$ description logics}
Temporalisations of DLs are known in the literature (see, e.g.,
\cite{ArtaleF00a,Bettini97a}); as well as spatialisations of DLs  (see, e.g.,
\cite{HaarslevLM99a}). The present work considers a spatio-temporalisation of
the well-known family $\alcd$ of DLs with a concrete domain \cite{BaaderH91a}.
Specifically, we consider, at the same time, a temporalisation of the roles of the family and a spatialisation of its concrete domain.
\subsection{Concrete domain}
\begin{definition}[concrete domain \cite{BaaderH91a}]\label{cddefinition}
A concrete domain $\textsl{D}$ consists of a pair
$(\Delta _{\textsl{D}},\Phi _{\textsl{D}})$, where
$\Delta _{\textsl{D}}$ is a set of (concrete) objects, and
$\Phi _{\textsl{D}}$ is a set of predicates over the objects in $\Delta _{\textsl{D}}$.
Each predicate $P\in\Phi _{\textsl{D}}$ is associated
with an arity $n$ and we have
$P\subseteq (\Delta _{\textsl{D}})^n$.
\end{definition}
\begin{definition}[admissibility \cite{BaaderH91a}]\label{cdadmissibility}
A concrete domain $\textsl{D}$ is admissible if:
(1) the set of its predicates is closed under negation and
    contains a predicate for $\Delta _{\textsl{D}}$; and
(2) the satisfiability problem for finite conjunctions of predicates is decidable.
\end{definition}
\subsection{The concrete domains $\textsl{D}_x$,
            with $x$ spatial RA}
Any spatial RA $x$ for
    which the atoms are Jointly Exhaustive and Pairwise Disjoint (henceforth JEPD), and
    such that the atomic relations form a decidable subclass, can be used to generate a
concrete domain $\textsl{D}_x$ for members of the family $\xdl$ of qualitative theories
for spatial change. Such a concrete domain is used for representing knowledge on
$p$-tuples of objects of the spatial domain at hand, $p$ being the arity of the $x$ relations;
stated otherwise, the $x$ relations will be used as the predicates of $\textsl{D}_x$.

\subsection{Admissibility of the concrete domains $\textsl{D}_x$,
            with $x\in\{\rcc8 ,\cdalg ,\atra\}$}\label{admissibility}
Let $x\in\{\rcc8 ,\cdalg ,\atra\}$. The concrete domain generated by $x$, $\textsl{D}_x$, can be written as
$\textsl{D}_x=(\Delta _{\textsl{D}_x},\Phi _{\textsl{D}_x})$, with:
                         $\textsl{D}_{\rcc8}   =(\rtopspace ,2^{\rccats})$,
                         $\textsl{D}_{\cdalg}  =(\deuxdp ,2^{\cdalgats})$ and
                         $\textsl{D}_{\atra}  =(\deuxdo ,2^{\atraats})$,
where
$\rtopspace$ is the set of regions of a topological space $\topspace$;
    $\deuxdp$ is the set of 2D points;
    $\deuxdo$ is the set of 2D orientations; and
$\xat$, as we have seen, is the set of $x$ atoms
    ---$2^{\xat}$ is thus the set of all $x$ relations.
%

Admissibility of the concrete domains $\textsl{D}_x$
is an immediate consequence of (decidability and) tractability of the subset
$\{\{r\}|r\in\xat\}$ of $x$ atomic relations, for each
$x\in\{\rcc8 ,\cdalg ,\atra\}$. The reader is referred to \cite{RenzN99a} for $x=\rcc8$, to \cite{Ligozat98a}
for $x=\cdalg$, and to \cite{IsliC98a,IsliC00a} for $x=\atra$:
\begin{theorem}\label{admissibilitythm}
Let $x\in\{\rcc8 ,\cdalg ,\atra\}$. The concrete domain $\textsl{D}_x$ is admissible. \cqfd
\end{theorem}
\subsection{Syntax of $\xdl$ concepts}
\begin{definition}[$\xdl$ concepts]\label{defxdlconcepts}
Let $x$ be an $\rcc8$-like $p$-ary spatial RA. Let $N_C$, $N_R$ and $N_{cF}$ be mutually disjoint and countably infinite sets of concept
names, role names, and concrete features, respectively; and $N_{aF}$ a countably infinite subset
of $N_R$ whose elements are abstract features. A (concrete)
feature chain is any finite composition $f_1\ldots f_ng$ of $n\geq 0$ abstract
features $f_1,\ldots ,f_n$ and one concrete feature $g$. The set of $\xdl$
concepts is the smallest set such that:
\begin{enumerate}
  \item\label{defxdlconceptsone} $\top$ and $\bot$ are $\xdl$ concepts
  \item\label{defxdlconceptstwo} an $\xdl$ concept name is an $\xdl$
    (atomic) concept
  \item\label{defxdlconceptsthree} if
    $C$ and $D$ are $\xdl$ concepts;
    $R$ is a role (in general, and an abstract feature in particular);
    $u_1,\ldots ,u_p$ are feature chains; and
    $P$ is an $\xdl$ predicate,
    then the following expressions are also $\xdl$ concepts:
    \begin{enumerate}
      \item\label{defxdlconceptsthreea} $\neg C$,
            $C\sqcap D$,
            $C\sqcup D$,
            $\exists R.C$,
            $\forall R.C$; and
      \item\label{defxdlconceptsthreeb}
            $\exists (u_1)\ldots (u_p).P$.
    \end{enumerate}
\end{enumerate}
\end{definition}
We denote by $\mtalc$ the sublanguage of $\xdl$ given by rules
\ref{defxdlconceptsone}, \ref{defxdlconceptstwo} and
\ref{defxdlconceptsthreea} in Definition \ref{defxdlconcepts},
which is the temporal component of $\xdl$. It is worth noting that $\mtalc$
does not consist of a mere temporalisation of $\alc$
\cite{Schmidt-SchaussS91a}. Indeed, $\alc$ contains only general roles,
whereas $\mtalc$ contains abstract features as well. A mere temporalisation of $\alc$ (i.e., $\mtalc$
without abstract features) cannot capture the expressiveness of
well-known modal temporal logics, including Propositional Linear Temporal Logic
$\pltl$, the computation tree logic $\ctl$, and the subsuming full branching modal temporal logic $\ctlstar$
\cite{Emerson90a}.
Given two integers $p\geq 0$ and $q\geq 0$, the sublanguage of
$\xdl$ (resp. $\mtalc$) whose concepts involve at most $p$
general roles, and $q$ abstract features
will be referred to as $\xdlpq$ (resp. $\mtalcpq$). The particuler case
$(p,q)=(0,q)$ with $q\geq 0$ is discussed in Section \ref{particularcases}, where we provide a
translation of $\ctlstar$ to $\mtalczq$.
\begin{definition}[subconcept]\label{defsubconcepts}
The set $\subc (C)$ of subconcepts of an $\xdl$ concept $C$ is defined inductively as follows:
\begin{enumerate}
  \item $\subc (\top )=\{\top\}$, $\subc (\bot )=\{\bot\}$
  \item $\subc (A)=\{A\}$, $\subc (\neg A)=\{\neg A\}$, for all atomic concepts $A$
  \item $\subc (C\sqcap D)=\{C\sqcap D\}\cup\subc (C)\cup\subc (D)$,
  \item $\subc (C\sqcup D)=\{C\sqcup D\}\cup\subc (C)\cup\subc (D)$,
  \item $\subc (\neg (C\sqcap D))=\{\neg (C\sqcap D)\}\cup\subc (\neg C)\cup\subc (\neg D)$,
  \item $\subc (\neg (C\sqcup D))=\{\neg (C\sqcup D)\}\cup\subc (\neg C)\cup\subc (\neg D)$,
  \item $\subc (\exists R.C)=\{\exists R.C\}\cup\subc (C)$,
  \item $\subc (\forall R.C)=\{\forall R.C\}\cup\subc (C)$,
  \item $\subc (\neg\exists R.C)=\subc (\forall R.\neg C)$,
  \item $\subc (\neg\forall R.C)=subc (\exists R.\neg C)$,
  \item $\subc (\exists (u_1)\ldots (u_p).P)=\{\exists (u_1)\ldots (u_p).P\}$.
  \item $\subc (\neg\exists (u_1)\ldots (u_p).P)=\{\exists (u_1)\ldots (u_p).\overline{P}\}$.
\end{enumerate}
\end{definition}
We now define weakly cyclic TBoxes.
\subsection{Weakly cyclic TBoxes}
An ($\xdl$ terminological) axiom is an expression of the form $A\doteq C$, $A$ being
a concept name and $C$ a concept. A TBox is a finite set of axioms, with
the condition that no concept name appears more than once as the left hand side of
an axiom.

Let $T$ be a TBox. $T$ contains two kinds of concept names: concept names appearing
as the left hand side of an axiom of $T$ are defined concepts; the others are
primitive concepts. A defined concept $A$ ``{\em directly uses}'' a defined
concept $B$ $\iff$ $B$ appears in the right hand side of the axiom defining $A$. If
``{\em uses}'' is the transitive closure of ``{\em directly uses}'' then $T$
contains a cycle $\iff$ there is a defined concept $A$ that ``{\em uses}'' itself.
$T$ is cyclic if it contains a cycle; it is acyclic otherwise. $T$ is
weakly cyclic if it satisfies the following two conditions:
\begin{enumerate}
  \item Whenever $A$ uses $B$ and $B$ uses $A$, we have $B=A$ ---the only possibility for a defined concept to get
involved in a cycle is to appear in the right hand side of the axiom
defining it.
  \item
	All possible occurrences of a defined concept $B$ in the right
    	hand side of the axiom $B\doteq C$ defining $B$ itself, are within the scope of exactly one quantifier
        (in other words, there is no free ocurrence of $B$ in C, and no occurrence of $B$ in $C$ is within the scope of
        more than one quantifier).
\end{enumerate}
\begin{definition}[depths of a defined concept]\label{depthsdefinition}
Let $B$ be a defined concept, and $C$ a concept. The set of depths of $B$ in $C$, $\profondeurs (B,C)$, is the set of
all integers $d$ such that $B$ has an occurrence in $C$ whithin the scope of $d$ quantifiers. $\profondeurs (B,C)$ is defined inductively as follows:
\begin{enumerate}
  \item\label{depthsun} if $B$ has no occurrence in $C$, $\profondeurs (B,C)=\emptyset$,
  \item $\profondeurs (B,B)=\{0\}$,
  \item $\profondeurs (B,\neg C)=\profondeurs (B,C)$,
  \item $\profondeurs (B,C\sqcap D)=\profondeurs (B,C\sqcup D)=\profondeurs (B,C)\cup\profondeurs (B,D)$,
  \item\label{depthsquatre} $\profondeurs (B,\exists R.C)=\profondeurs (B,\forall R.C)=\{d+1:d\in\profondeurs (B,C)\}$
\end{enumerate}
\end{definition}
\begin{remark}
In Definition \ref{depthsdefinition}:
\begin{enumerate}
  \item Item (\ref{depthsun}) includes the following particular case: $\profondeurs (B,\exists (u_1)\ldots (u_p).P)=\emptyset$; and
  \item Item (\ref{depthsquatre}) includes the particular case:  if  $\profondeurs (B,C)=\emptyset$ then $\profondeurs (B,\exists R.C)=\profondeurs (B,\forall R.C)=\emptyset$.
\end{enumerate}
\end{remark}

A weakly cyclic TBox can now be defined formally as follows:
\begin{definition}[weakly cyclic TBox]\label{defwctb}
A TBox $T$ is weakly cyclic if and only if it satisfies what follows:
\begin{enumerate}
  \item whenever two defined concepts $A$ and $B$ are such that $A$ uses $B$ and $B$ uses $A$, we have $B=A$; and
  \item all axioms $B\doteq C$ of $T$ verify the following: $\profondeurs (B,C)=\emptyset$ or $\profondeurs (B,C)=\{1\}$.
\end{enumerate}
\end{definition}
\begin{definition}Let $T$ be a weakly cyclic TBox.
\begin{enumerate}
  \item An axiom $B\doteq C$ of $T$ is cyclic if $\profondeurs (B,C)=\{1\}$; it is acyclic otherwise
  \item A defined concept $B$ of $T$ is cyclic if the axiom $B\doteq C$ defining it is cyclic; it is acyclic otherwise
  \item A cyclic axiom of $T$ is said to be a necessity axiom if it is of either of the following forms:
    \begin{enumerate}
      \item $B\doteq C\sqcap\forall R.B$ where $R$ is a role, either general or functional; and $C$ a concept such that $\profondeurs (B,C)=\emptyset$
      \item $B\doteq C_1\sqcap (C_2\sqcup\forall R.B)$ where $R$ is a role, either general or functional; and $C_1$ and $C_2$ concepts such that $\profondeurs (B,C_1)=\profondeurs (B,C_2)=\emptyset$
    \end{enumerate}
  \item A cyclic axiom of $T$ is said to be an eventuality axiom if it is of either of the following forms:
    \begin{enumerate}
      \item $B\doteq C\sqcup\exists R.B$ where $R$ is a role, either general or functional; and $C$ a concept such that $\profondeurs (B,C)=\emptyset$
      \item $B\doteq C_1\sqcup (C_2\sqcap\exists R.B)$ where $R$ is a role, either general or functional; and $C_1$ and $C_2$ concepts such that $\profondeurs (B,C_1)=\profondeurs (B,C_2)=\emptyset$
    \end{enumerate}
  \item A defined concept of $T$ is a necessity defined concept if the axiom defining it is a necessity axiom
  \item A defined concept of $T$ is an eventuality defined concept if the axiom defining it is an eventuality axiom
  \item The necessity defined concept $B_1$ and the eventuality defined concept $B_2$ defined, respectively, by the axioms $B_1\doteq C\sqcap\forall R.B_1$ and $B_2\doteq\neg C\sqcup\exists R.B_1$
           are each other's duals
  \item The necessity defined concept $B_1$ and the eventuality defined concept $B_2$ defined, respectively, by the axioms $B_1\doteq C_1\sqcap (C_2\sqcup \forall R.B_1)$ and $B_2\doteq\neg C_1\sqcup (\neg C_2\sqcap \exists R.B_2)$
           are each other's duals
\end{enumerate}
\end{definition}
From now on, we restrict ourselves, exclusively, to weakly cyclic TBoxes $T$ such that
\begin{enumerate}
  \item for all necessity or eventuality defined concepts $B$ of $T$, $T$ also has the defined concept consisting of the dual of $B$; and
  \item all defined concepts $B$ verify the following:
    \begin{enumerate}
      \item $B$ is acyclic,
      \item $B$ is a necessity defined concept, or
      \item $B$ is an eventuality defined concept
    \end{enumerate}
\end{enumerate}

In the rest of the paper, unless explicitly stated otherwise, we denote
concepts reducing to concept names by the letters $A$ and $B$,
possibly complex concepts by the letters $C$, $D$, $E$,
general roles by the letter $R$,
abstract features by the letter $f$, concrete features by the letters $g$ and $h$,
feature chains by the letter $u$,
(possibly complex) predicates by the letter $P$.

\begin{example}
Due to  lack of space, an example supposed to come here is added as additional material, as a separate file including a brief
background on the ternary spatial RA $\atra$ \cite{IsliC98a,IsliC00a} and an illustration of the use of $\xdlzoatra$ in robot navigation.
\end{example}
\section{Semantics of $\xdl$}
Let $\textsl{D}_x$ be an admissible spatial concrete domain generated by a $p$-ary spatial RA $x$.
$\xdl$ concepts will be interpreted over $k$-ary $\Sigma$-trees.
\begin{definition}[$k$-ary $\Sigma$-tree]\label{karymtree}
Let $\Sigma$ and $K=\{d_1,\ldots ,d_k\}$, $k\geq 1$, be two disjoint alphabets: $\Sigma$ is a labelling alphabet and $K$ an
alphabet of directions. A (full) $k$-ary tree is an infinite tree
whose nodes $\alpha\in K^*$ have exactly $k$ immediate successors each,
$\alpha d_1,\ldots ,\alpha d_k$. A $\Sigma$-tree is a tree whose nodes are
labelled with elements of $\Sigma$. A (full) $k$-ary $\Sigma$-tree is
a $k$-ary tree $t$ which is also a $\Sigma$-tree, which we consider as
a mapping $t:K^*\rightarrow\Sigma$ associating with each node
$\alpha\in K^*$ an element $t(\alpha )\in\Sigma$. The empty word, $\epsilon$, denotes
the root of $t$. Given a node $\alpha\in K^*$ and a direction $d\in K$,
the concatenation of $\alpha$ and $d$, $\alpha d$, denotes the $d$-successor of
$\alpha$. The level $|\alpha |$ of a node $\alpha$ is the length of $\alpha$ as a word. We can
thus think of the edges of $t$ as being labelled with directions from $K$, and of the nodes of
$t$ as being labelled with letters from $\Sigma$. A partial $k$-ary
$\Sigma$-tree (over the set $K$ of directions) is a $\Sigma$-tree with
the property that a node may not have a $d$-successor for each
direction $d$; in other terms, a partial $k$-ary $\Sigma$-tree is a
$\Sigma$-tree which is a prefix-closed\footnote{$t$ is prefix-closed
  if, for all nodes $\alpha$, if $t$ is defined for $\alpha$ then it
is defined for all nodes $\alpha '$ consisting of prefixes of
  $\alpha$.} partial function $t:K^*\rightarrow\Sigma$.
\end{definition}
$\xdl$ is equipped with a Tarski-style possible worlds
semantics. $\xdl$ interpretations are spatio-temporal structures
consisting of $k$-ary trees $t$, representing
\mbox{$k$-immediate-successor} branching time, together with an interpretation
function associating with each primitive concept $A$ the nodes of $t$
at which $A$ is true, and, additionally, associating with each
concrete feature $g$ and each node $u$ of $t$, the value at $u$ (seen as
a time instant) of the spatial concrete object referred to by $g$. Formally:
\begin{definition}[interpretation]
Let $x$ be an $\rcc8$-like $p$-ary spatial RA and $K=\{d_1,\ldots ,d_k\}$ a set
of $k$ directions.
An interpretation $\textsl{I}$ of $\xdl$ consists of a pair $\textsl{I}=(t _{\textsl{I}},.^{\textsl{I}})$, where
$t _{\textsl{I}}$ is a $k$-ary tree and $.^{\textsl{I}}$ is an
interpretation function mapping
each primitive concept $A$ to a subset $A^{\textsl{I}}$ of $K^*$;
each role $R$ to a subset $R^{\textsl{I}}$ of $\{(u,ud)\in K^*\times K^*:\mbox{ }d\in K\}$,
so that $R^{\textsl{I}}$ is functional if $R$ is an abstract feature; and
each concrete feature $g$ to a total function $g^{\textsl{I}}$ 
from $K^*$ onto the set $\Delta _{\textsl{D}_x}$ of (concrete) objects of the concrete
domain $\textsl{D}_x$.
\end{definition}
Given an $\xdl$ interpretation $\textsl{I}=(t_{\textsl{I}},.^{\textsl{I}})$, a feature chain
$u=f_1\ldots f_ng$, and a node $v_1$, we denote by $u^{\textsl{I}}(v_1)$
the value $g^{\textsl{I}}(v_2)$, where $v_2$ is the $f_1^{\textsl{I}}\ldots f_n^{\textsl{I}}$-successor of $v_1$; i.e., $v_2$ is so that there exists a sequence
$v_1=w_0,w_1,\ldots ,w_n=v_2$ verifying $(w_i,w_{i+1})\in
f_{i+1}^{\textsl{I}}$,
for all $i\in\{0,\ldots ,n-1\}$ (in other words,
     $v_2=(f_n^{\textsl{I}}\circ\ldots\circ f_1^{\textsl{I}})(v_1)
            =f_n^{\textsl{I}}(\ldots (f_1^{\textsl{I}}(v_1))\ldots )$).
\begin{definition}[satisfiability $\wrt$ a TBox]
Let $x$ be an $\rcc8$-like $p$-ary spatial RA, $K=\{d_1,\ldots ,d_k\}$ a set
of $k$ directions, $C$ an $\xdl$ concept, $\textsl{T}$ an $\xdl$
weakly cyclic TBox, and $\textsl{I}=(t_{\textsl{I}},.^{\textsl{I}})$ an
$\xdl$ interpretation. The satisfiability, by a node $s$ of $t_{\textsl{I}}$, of $C$ $\wrt$ to $\textsl{T}$, denoted $\textsl{I},s\models\langle C,\textsl{T}\rangle$, is defined inductively as follows:
\begin{enumerate}
  \item $\textsl{I},s\models\langle\top ,\textsl{T}\rangle$
  \item $\textsl{I},s\not\models\langle\bot ,\textsl{T}\rangle$
  \item For all primitive concepts $A$:
    \begin{enumerate}
      \item $\textsl{I},s\models\langle A,\textsl{T}\rangle$ $\iff$ $s\in A^{\textsl{I}}$
       \item $\textsl{I},s\models\langle\neg A,\textsl{T}\rangle$ $\iff$ $s\notin A^{\textsl{I}}$
    \end{enumerate}
  \item $\textsl{I},s\models\langle B,\textsl{T}\rangle$ $\iff$
    $\textsl{I},s\models\langle C,\textsl{T}\rangle$,
           $\textsl{I},s\models\langle\neg B,\textsl{T}\rangle$ $\iff$
    $\textsl{I},s\models\langle\neg C,\textsl{T}\rangle$, for all defined
    concepts $B$ defined by the axiom $B\doteq C$ of $\textsl{T}$,
    such that $B$ does not occur in $C$, the right
    hand side of the axiom (in other words, such that $\profondeurs (B,C)=\emptyset$).
  \item for all eventuality defined concepts $B$ defined by the axiom $B\doteq C\sqcup\exists R.B$,
	$\textsl{I},s\models\langle B,\textsl{T}\rangle$ $\iff$ there exists $s_0=s,\ldots ,s_i$, with $i\geq 0$, such that:
    \begin{enumerate}
      \item $(s_j,s_{j+1})\in R^{\textsl{I}}$, for all $j$ such that $0\leq j<i$; and
      \item $\textsl{I},s_i\models\langle C,\textsl{T}\rangle$
    \end{enumerate}
  \item for all eventuality defined concepts $B$ defined by the axiom $B\doteq C_1\sqcup (C_2\sqcap\exists R.B)$,
	$\textsl{I},s\models\langle B,\textsl{T}\rangle$ $\iff$ there exists $s_0=s,\ldots ,s_i$, with $i\geq 0$, such that:
    \begin{enumerate}
      \item $(s_j,s_{j+1})\in R^{\textsl{I}}$, for all $j$ such that $0\leq j<i$;
      \item $\textsl{I},s_j\models\langle C_2,\textsl{T}\rangle$, for all $j$ such that $0\leq j<i$; and
      \item $\textsl{I},s_i\models\langle C_1,\textsl{T}\rangle$
    \end{enumerate}

  \item for all necessity defined concepts $B$ defined by the axiom $B\doteq C\sqcap\forall R.B$,
	$\textsl{I},s\models\langle B,\textsl{T}\rangle$ $\iff$
    \begin{enumerate}
      \item $\textsl{I},s\models\langle C,\textsl{T}\rangle$; and
      \item $\textsl{I},s'\models\langle B,\textsl{T}\rangle$, for all $s'$ such that $(s,s')\in R^{\textsl{I}}$
    \end{enumerate}
  \item for all necessity defined concepts $B$ defined by the axiom $B\doteq C_1\sqcap (C_2\sqcup\forall R.B)$,
	$\textsl{I},s\models\langle B,\textsl{T}\rangle$ $\iff$
    \begin{enumerate}
      \item $\textsl{I},s\models\langle C_1,\textsl{T}\rangle$; and
      \item $\textsl{I},s\models\langle C_1,\textsl{T}\rangle$ or $\textsl{I},s'\models\langle B,\textsl{T}\rangle$, for all $s'$ such that $(s,s')\in R^{\textsl{I}}$
    \end{enumerate}

  \item $\textsl{I},s\models\langle C\sqcap D,\textsl{T}\rangle$ $\iff$
    $\textsl{I},s\models\langle C,\textsl{T}\rangle$ and $\textsl{I},s\models\langle D,\textsl{T}\rangle$
  \item $\textsl{I},s\models\langle C\sqcup D,\textsl{T}\rangle$ $\iff$
    $\textsl{I},s\models\langle C,\textsl{T}\rangle$ or $\textsl{I},s\models\langle D,\textsl{T}\rangle$
  \item $\textsl{I},s\models\langle \neg B_1,\textsl{T}\rangle$ $\iff$
    $\textsl{I},s\models\langle B_2,\textsl{T}\rangle$, for all necessity or eventuality defined concepts $B_1$ whose dual is $B_2$
  \item $\textsl{I},s\models\langle \neg (C\sqcap D),\textsl{T}\rangle$ $\iff$
    $\textsl{I},s\models\langle \neg C,\textsl{T}\rangle$ or $\textsl{I},s\models\langle \neg D,\textsl{T}\rangle$
  \item $\textsl{I},s\models\langle \neg (C\sqcup D),\textsl{T}\rangle$ $\iff$
    $\textsl{I},s\models\langle \neg C,\textsl{T}\rangle$ and $\textsl{I},s\models\langle \neg D,\textsl{T}\rangle$

  \item $\textsl{I},s\models\langle\exists R.C,\textsl{T}\rangle$ $\iff$
    $\textsl{I},s'\models\langle C,\textsl{T}\rangle$, for some $s'$ such
    that $(s,s')\in R^{\textsl{I}}$
  \item $\textsl{I},s\models\langle\forall R.C,\textsl{T}\rangle$ $\iff$
    $\textsl{I},s'\models\langle C,\textsl{T}\rangle$, for all $s'$ such
    that $(s,s')\in R^{\textsl{I}}$

  \item $\textsl{I},s\models\langle\neg\exists R.C,\textsl{T}\rangle$ $\iff$
    $\textsl{I},s\models\langle\forall R.\neg C,\textsl{T}\rangle$
  \item $\textsl{I},s\models\langle\neg\forall R.C,\textsl{T}\rangle$ $\iff$
    $\textsl{I},s\models\langle\exists R.\neg C,\textsl{T}\rangle$

  \item $\textsl{I},s\models\langle\exists (u_1)\ldots (u_p).P,\textsl{T}\rangle$ $\iff$
    $P(u_1^{\textsl{I}}(s),\ldots ,u_p^{\textsl{I}}(s))$,
           $\textsl{I},s\models\langle\neg\exists (u_1)\ldots (u_p).P,\textsl{T}\rangle$ $\iff$
    $\overline{P}(u_1^{\textsl{I}}(s),\ldots ,u_p^{\textsl{I}}(s))$
\end{enumerate}
A concept $C$ is satisfiable $\wrt$ a TBox $\textsl{T}$ 
$\iff$ $\textsl{I},s\models\langle C,\textsl{T}\rangle$, for some $\xdl$
interpretation $\textsl{I}$, and some state
$s\in t_{\textsl{I}}$, in which case the pair $(\textsl{I},s)$ is a model
of $C$ $\wrt$ $\textsl{T}$; $C$ is insatisfiable (has no models) $\wrt$ $\textsl{T}$,
otherwise. $C$ is valid $\wrt$ $\textsl{T}$ $\iff$ the negation, $\neg
C$, of $C$ is insatisfiable $\wrt$ $\textsl{T}$.
\end{definition}
\section{The satisfiability of an $\xdl$ concept $\wrt$ a weakly cyclic TBox}
Let $C$ be an $\xdl$ concept and $\textsl{T}$ an $\xdl$ weakly cyclic TBox. We define $\textsl{T}\oplus C$ as the TBox
$\textsl{T}$ augmented with the axiom 
$B_{init}\doteq C$, $B_{init}$ being a fresh defined concept (not occurring in $\textsl{T}$):
\begin{footnotesize}
\begin{eqnarray}
\textsl{T}\oplus C=\textsl{T}\cup\{B_{init}\doteq C\}\nonumber
\end{eqnarray}
\end{footnotesize}
In the sequel, we refer to $\textsl{T}\oplus C$ as the TBox $\textsl{T}$
augmented with $C$.
The idea now is that, satisfiability of $C$ $\wrt$ $\textsl{T}$ has (almost) been reduced to the emptiness
problem of $\textsl{T}\oplus C$, seen as a weak alternating automaton on
$k$-ary $\Sigma$-trees, for some labelling alphabet $\Sigma$ to be
defined later, with the
defined concepts as the states of
the automaton, $B_i$ as the initial state of the automaton, the axioms
as defining the transition function,
with the accepting condition derived from those defined concepts that are not eventuality concepts, and with $k$
standing for the number of concepts of the form $\exists R.D$ in a certain closure, to be defined later, of
$\textsl{T}\oplus C$.
\subsection{The Disjunctive Normal Form}
The notion of Disjunctive Normal Form (DNF) of a concept $C$ $\wrt$ to a TBox $\textsl{T}$, $\dnfone (C,\textsl{T})$,
is crucial for the rest of the paper.
Such a form results, among other things, from the use of De Morgan's Laws to decompose a concept so that, in the
final form, the negation symbol outside the scope of a (existential or universal) quantifier occurs only in front
of primitive concepts.

Given a (concrete) feature chain $u$, we define $Exists(u)$ as follows:\\
$
Exists(u)=
  \left\{
                   \begin{array}{l}
                         \emptyset\mbox{$\;\;\;$ if $u$ reduces to a concrete feature},  \\
                         \{\exists f_1.(\exists f_2.(\cdots .(\exists f_n.\top)\cdots ))\}  \\
                         \mbox{$\;\;\;\;\;$ otherwise ($u$ of the form $f_1f_2\cdots f_ng$)}  \\
                   \end{array}
  \right.
$

\begin{definition}[first DNF]\label{firstdnf}
The first Disjunctive Normal Form ($\dnfone$) of an $\xdl$ concept $C$ $\wrt$ an $\xdl$ weakly cyclic TBox $\textsl{T}$,
$\dnfone (C,\textsl{T})$, is defined inductively as follows:
\begin{enumerate}
  \item for all primitive concepts $A$: $\dnfone (A,\textsl{T})=\{\{A\}\}$, $\dnfone (\neg A,\textsl{T})=\{\{\neg A\}\}$
  \item $\dnfone (\top ,\textsl{T})=\{\emptyset\}$, $\dnfone (\bot ,\textsl{T})=\emptyset$
  \item for all acyclic defined concepts $B$: $\dnfone (B,\textsl{T})=\dnfone (E,\textsl{T})$,
    $\dnfone (\neg B,\textsl{T})=\dnfone (\neg E,\textsl{T})$, where $E$ is the right hand side of the axiom $B\doteq E$
    defining $B$
%

  \item for all eventuality defined concepts $B$ defined by the axiom $B\doteq C\sqcup\exists R.B$,
	$\dnfone (B,\textsl{T})=\dnfone (C,\textsl{T})\cup\{\{\exists R.B\}\}$
  \item for all eventuality defined concepts $B$ defined by the axiom $B\doteq C_1\sqcup (C_2\sqcap\exists R.B)$,
	$\dnfone (B,\textsl{T})=\dnfone (C_1,\textsl{T})\cup\prod (\dnfone (C_2,\textsl{T}),\{\{\exists R.B\}\})$
  \item for all necessity defined concepts $B$ defined by the axiom $B\doteq C\sqcap\forall R.B$,
	$\dnfone (B,\textsl{T})=\prod (\dnfone (C,\textsl{T}),\{\{\forall R.B\}\})$
  \item for all necessity defined concepts $B$ defined by the axiom $B\doteq C_1\sqcap (C_2\sqcup\forall R.B)$,
	$\dnfone (B,\textsl{T})=\prod (\dnfone (C_1,\textsl{T}),\dnfone (C_2,\textsl{T})\cup\{\{\forall R.B\}\})$
  \item for all necessity or eventuality defined concepts $B_1$ whose dual is the defined concept $B_2$,
	$\dnfone (\neg B_1,\textsl{T})=\dnfone (B_2,\textsl{T})$
  \item $\dnfone (C\sqcap D,\textsl{T})=\prod (\dnfone (C,\textsl{T}),\dnfone (D,\textsl{T}))$
  \item $\dnfone (C\sqcup D,\textsl{T})=\dnfone (C,\textsl{T})\cup\dnfone (D,\textsl{T})$
  \item $\dnfone (\exists R.C,\textsl{T})=\{\{\exists R.C\}\}$
  \item $\dnfone (\forall R.C,\textsl{T})=\{\{\forall R.C\}\}$
  \item $\dnfone (\exists (u_1)\ldots (u_p).P,\textsl{T})=\{\{\exists (u_1)\ldots (u_p).P\}\cup Exists(u_1)\cup\ldots\cup Exists(u_p)$
  \item $\dnfone (\neg (C\sqcap D),\textsl{T})=\dnfone (\neg C,\textsl{T})\cup\dnfone (\neg D,\textsl{T})$
  \item $\dnfone (\neg (C\sqcup D),\textsl{T})=\prod (\dnfone (\neg C,\textsl{T}),\dnfone (\neg D,\textsl{T}))$
  \item $\dnfone (\neg\exists R.C,\textsl{T})=\{\{\forall R.\neg C\}\}$
  \item $\dnfone (\neg\forall R.C,\textsl{T})=\{\{\exists R.\neg C\}\}$
  \item $\dnfone (\neg\exists (u_1)\ldots (u_p).P,\textsl{T})=\{\{\exists (u_1)\ldots (u_p).\overline{P}\}\cup Exists(u_1)\cup\ldots\cup Exists(u_p)$
\end{enumerate}
where $\prod$ is defined as follows:
\begin{enumerate}
  \item 
$
\prod (\{S\},\{T\})=
  \left\{
                   \begin{array}{l}
                         \emptyset\mbox{ if }\{A,\neg A\}\subseteq S\cup T\mbox{ for some }  \\
                         \mbox{$\;\;\;\;\;\;\;\;\;\;\;\;\;\;\;\;$ primitive concept }A,  \\
                         \{S\cup T\}\mbox{ otherwise}  \\
                   \end{array}
  \right.
$
  \item $\prod (\{S_1,\ldots ,S_n\},\{T_1,\ldots ,T_m\})=\displaystyle
    \bigcup _{i\in\{1,\ldots ,n\},j\in\{1,\ldots ,m\}}\prod (\{S_i\},\{T_j\})$
\end{enumerate}
\end{definition}
Note that the $\dnfone$ function checks satisfiability at the propositional level,
in the sense that, given a concept $C$, $\dnfone (C,\textsl{T})$ is either empty, or is such
that for all $S\in\dnfone (C,\textsl{T})$, $S$ does not contain both $A$ and $\neg A$, $A$
being a primitive concept. Furthermore, given a set $S\in\dnfone (C,\textsl{T})$, all
elements of $S$ are concepts of either of the following forms:
$A$ or $\neg A$, where $A$ is a primitive concept;
$\forall R.D$; or
$\exists (u_1)\ldots (u_p).P$.
%
\begin{definition}[the $\pceaone$ partition]
Let $C$ be an $\xdl$ concept, $\textsl{T}$ an $\xdl$ TBox, $S\in\dnfone (C,\textsl{T})$ and $N_{aF}^*$ the
language of all finite words over the alphabet $N_{aF}$. The $\pcea$ partition of $S$, $\pceaone (S)$, is
defined as
$\pceaone (S)=\{S_{prop},S_{csp},S_{\exists}\}$, where:
\begin{eqnarray}
  S_{prop}&=\{A:A\in S\mbox{ and $A$ primitive concept}\}  \nonumber  \\
                 &\cup\{\neg A:\neg A\in S\mbox{ and $A$ primitive concept}\}  \nonumber  \\
  S_{csp}&=
                   \{\exists (u_1)\ldots (u_p).P:\exists (u_1)\ldots (u_p).P\in S\}
  \nonumber
\end{eqnarray}

and $S_{\exists}$ is computed as follows :
\begin{enumerate}
  \item Initialise $S_{\exists}$ to the empty set : $S_{\exists}=\emptyset$
  \item For all $\exists R.C$ in $S$ with $R$ general role:
						$S_{\exists}=S_{\exists}\cup\{\exists R.(C\sqcap C_1\cdots\sqcap C_k): \{C_1,...,C_k\}=\{D:\forall R.D\in S\}\}$
  \item For all abstract features $f$ such that $S$ contains elements of the form $\exists f.C$:
						$S_{\exists}=S_{\exists}\cup\{\exists f.(C_1\sqcap\cdots\sqcap C_k\sqcap D_1\cdots\sqcap D_l): \{C_1,...,C_k\}=\{E:\exists f.E\in S\}\mbox{ and }\{D_1,...,D_l\}=\{E:\forall f.E\in S\}\}$
\end{enumerate}

\end{definition}
The second $\dnf$ of a concept $C$ $\wrt$ a TBox $\textsl{T}$, $\dnftwo (C,\textsl{T})$, is now
introduced. This consists of the $\dnfone$ of $C$ $\wrt$ $\textsl{T}$, $\dnfone (C,\textsl{T})$, as
given by Definition \ref{firstdnf}, in which each element $S$ is replaced with $S^f=S_{prop}\cup S_{csp}\cup S_{\exists}$. Formally:
\begin{definition}[second DNF]
Let $x\in\{\rcc8 ,\cdalg ,\atra\}$, $C$ be an $\xdl$ concept, and $\textsl{T}$ an $\xdl$ weakly cyclic TBox. The
second Disjunctive Normal Form ($\dnf2$) of $C$ $\wrt$ $\textsl{T}$, $\dnftwo (C,\textsl{T})$, is
defined as $\dnftwo (C,\textsl{T})=\{S^f:S\in\dnfone (C,\textsl{T})\}$.
\end{definition}
Given an $\xdl$ concept $C$ and an $\xdl$ TBox $\textsl{T}$, we can now
use the second DNF, $\dnftwo$, to define the closure $(\tpc )^*$ of
$\tpc$, the TBox $T$ augmented with $C$.
\begin{definition}[closure of $\tpc$] Let $C$ be an $\xdl$ concept and $T$ an
  $\xdl$ weakly cyclic TBox. The closure $(\tpc )^*$ of $\tpc$ is defined by the procedure
of Figure \ref{dnf2procedure}, which also outputs a partial order $\partialorder$
on the defined concepts of $(\tpc )^*$.
\end{definition}
\begin{remark}
The axioms of $(\tpc )^*$ are of the form $B=\{S_1,\ldots ,S_m\}$; for all $S\in\{S_1,\ldots ,S_m\}$, all elements of $S$
are of either of the following forms:
\begin{enumerate}
  \item $A$ or $\neg A$, where $A$ is a primitive concept;
  \item $\exists R.(B_1\sqcap\cdots\sqcap B_k)$, $R$ being a general role or an abstract feature, and $B_j$ a defined concept, for all $j\in\{1,\ldots ,k\}$; or
  \item $\exists (u_1)\ldots (u_p).P$.
\end{enumerate}
\end{remark}
We also need the closure of a concept $C$ $\wrt$ a TBox $\textsl{T}$,
$\closure (C,\textsl{T})$, which is defined as the union of the right hand sides of the axioms in $(\tpc )^*$. Formally:
\begin{definition}[closure of a concept $\wrt$ a TBox]
The closure of an $\xdl$ concept $C$ $\wrt$ an $\xdl$ TBox $\textsl{T}$,
$\closure (C,\textsl{T})$, is defined as follows:
\begin{footnotesize}
\begin{eqnarray}
\closure (C,\textsl{T})&=&\displaystyle\bigcup _{B\doteq E\mbox{ axiom of }(\tpc )^*}E\nonumber
\end{eqnarray}
\end{footnotesize}
\end{definition}
\begin{figure}
\begin{scriptsize}
\begin{enumerate}
  \item Initialise $\tpcf$ to $\tpc$: $\tpcf\leftarrow\tpc$;
  \item Initially, no defined concept of $\tpcf$ is marked;
  \item {\em while}($\tpcf$ contains defined concepts that are not marked)\{
    \begin{enumerate}
      \item consider a non marked defined concept $B_1$ from $\tpcf$;
      \item let $B_1\doteq E$ be the axiom from $\tpcf$ defining $B_1$;
      \item mark $B_1$;
      \item $\partialorder (B_1)=\emptyset$
      \item compute $\dnfone (E,\tpcf )$;
      \item $U_1=\emptyset$
      \item {\em for} all $S\in\dnfone (E,\tpcf )$
        \begin{enumerate}
          \item $U_2=S_{prop}\cup S_{csp}$
          \item $U_3=\emptyset$
          \item {\em for} all $\exists R.D\in S$ with $R$ general role$\{$
            \begin{enumerate}
              \item {\em if} $D$ is a defined concept of $\tpcf$ {\em then} $U_4=\{D\}$
              \item {\em else}
              \item[] \hskip 0.2cm {\em if}($\tpcf$ contains an axiom of the form $B_2\doteq D$) {\em then}
              \item[] \hskip 0.4cm $U_4=\{B_2\}$ ;
              \item[] \hskip 0.2cm {\em else}\{
              \item[] \hskip 0.4cm let $B_2$ be a fresh defined concept;
              \item[] \hskip 0.4cm add the axiom $B_2\doteq D$ to $\tpcf$:
              \item[] \hskip 0.8cm  $\tpcf\leftarrow\tpcf\cup\{B_2\doteq D\}$;
              \item[] \hskip 0.4cm $U_4=\{B_2\}$;
              \item[] \hskip 0.4cm $\}$
              \item {\em for} all $E$ such that $\forall R.E\in S\{$
              \item[] \hskip 0.2cm {\em if} $E$ is a defined concept of $\tpcf$ {\em then} $U_4=U_4\cup\{E\}$
              \item[] \hskip 0.2cm {\em else}
              \item[] \hskip 0.4cm {\em if}($\tpcf$ contains an axiom of the form $B_2\doteq E$) {\em then}
              \item[] \hskip 0.6cm $U_4=U_4\cup\{B_2\}$ ;
              \item[] \hskip 0.4cm {\em else}\{
              \item[] \hskip 0.6cm let $B_2$ be a fresh defined concept;
              \item[] \hskip 0.6cm add the axiom $B_2\doteq E$ to $\tpcf$:
              \item[] \hskip 1cm  $\tpcf\leftarrow\tpcf\cup\{B_2\doteq E\}$;
              \item[] \hskip 0.6cm $U_4=U_4\cup\{B_2\}$;
          \item[] \hskip 0.6cm \}
              \item $U_3=U_3\cup\{\exists R.(F_1\sqcap\cdots\sqcap F_\ell):\mbox{ }\{F_1,\cdots ,F_\ell\}=U_4\}$;
              \item $\partialorder (B_1)=\partialorder (B_1)\cup U_4$
          \end{enumerate}
          \item {\em for} all abstract features $f$ such that $S$ contains elements of the form $\exists f.C\{$
          \begin{enumerate}
              \item $U_4=\emptyset$
              \item {\em for} all $E$ such that $\exists f.E\in S$ or $\forall f.E\in S\{$
              \item[] \hskip 0.2cm {\em if} $E$ is a defined concept of $\tpcf$ {\em then} $U_4=U_4\cup\{E\}$
              \item[] \hskip 0.2cm {\em else}
              \item[] \hskip 0.4cm {\em if}($\tpcf$ contains an axiom of the form $B_2\doteq E$) {\em then}
              \item[] \hskip 0.6cm $U_4=U_4\cup\{B_2\}$ ;
              \item[] \hskip 0.4cm {\em else}\{
              \item[] \hskip 0.6cm let $B_2$ be a fresh defined concept;
              \item[] \hskip 0.6cm add the axiom $B_2\doteq E$ to $\tpcf$:
              \item[] \hskip 1cm  $\tpcf\leftarrow\tpcf\cup\{B_2\doteq E\}$;
              \item[] \hskip 0.6cm $U_4=U_4\cup\{B_2\}$;
          \item[] \hskip 0.6cm \}
              \item $U_3=U_3\cup\{\exists f.(F_1\sqcap\cdots\sqcap F_\ell):\mbox{ }\{F_1,\cdots ,F_\ell\}=U_4\}$;
              \item $\partialorder (B_1)=\partialorder (B_1)\cup U_4$
          \end{enumerate}
      \item $U_2=U_2\cup U_3$;
      \item $U_1=U_1\cup\{U_2\}$
        \end{enumerate}
      \item replace, in $\tpcf$, the axiom $B_1\doteq E$ with  the axiom $B_1\doteq U_1$;
      \item[\}]
    \end{enumerate}
\end{enumerate}
\caption{The closure $\tpcf$ of a weakly cyclic TBox $T$ augmented with a concept $C$, $\tpc$; and a partial order $\partialorder$ on the defined concepts of $\tpcf$.}\label{dnf2procedure}
\end{scriptsize}
\end{figure}
\begin{definition}
Let $C$ be an $\xdl$ concept and $\textsl{T}$ an $\xdl$ TBox. We denote by:
\begin{enumerate}
  \item $\concretefeatures (S)$, where $S\in\closure (C,\textsl{T})$, the set of concrete features of $S$:
    \begin{enumerate}
      \item[] in other words, $\concretefeatures (S)$ is the set of concrete features $g$ for which there exists a feature chain $u$ suffixed by $g$, such that $S$
           contains a predicate concept $\exists (u_1)\ldots (u_p).P$, with $u\in\{u_1,\ldots ,u_p\}$.
    \end{enumerate}
  \item $\concretefeatures (C,\textsl{T})=\displaystyle\bigcup _{S\in\closure (C,\textsl{T})}\concretefeatures (S)$,
    the set of concrete features of $C$ $\wrt$ $\textsl{T}$;
  \item $\ncf (C,\textsl{T})=|\concretefeatures (C,\textsl{T})|$,
    the number of concrete features of $C$ $\wrt$ $\textsl{T}$;
  \item $\abstractfeatures (C,\textsl{T})=\{f\in N_{aF}:\mbox{ for some concept $E$ there exists $S\in\closure (C,\textsl{T})$ s.t. }\exists f.E\in S\}$,
    the set of abstract features of $C$ $\wrt$ $\textsl{T}$;
  \item $\naf (C,\textsl{T})=|\abstractfeatures (C,\textsl{T})|$,
    the number of abstract features of $C$ $\wrt$ $\textsl{T}$;
  \item $\primitiveconcepts (C,\textsl{T})=\{A:\exists S\in\closure (C,\textsl{T})\mbox{ s. t. }\{A,\neg A\}\cap S_{prop}\not =\emptyset\}$,
    the set of primitive concepts of $C$ $\wrt$ $\textsl{T}$;
  \item $\definedconcepts (C,\textsl{T})$ is the set of defined concepts 
    in $(\tpc )^*$;
  \item $\reconcepts (C,\textsl{T})$,
    the set of relational existential (sub)concepts of $C$ $\wrt$ $\textsl{T}$, is
    the set of all $\exists R.D$ such that $R$ is a general role and there exists an axiom $B\doteq E$
    in $(\tpc )^*$ and $S$ in $E$ so that $\exists R.D\in S$;
  \item $\fbranchingfactor (C,\textsl{T})=\naf (C,\textsl{T})$, the functional branching factor of $C$ $\wrt$ $\textsl{T}$;
  \item $\rbranchingfactor (C,\textsl{T})=|\reconcepts (C,\textsl{T})|$,
    the relational branching factor of $C$ $\wrt$ $\textsl{T}$;
  \item $\branchingfactor (C,\textsl{T})=\fbranchingfactor (C,\textsl{T})+\rbranchingfactor (C,\textsl{T})$,
    the branching factor of $C$ $\wrt$ $\textsl{T}$.
\end{enumerate}
\end{definition}
We suppose that the relational existential concepts in $\reconcepts (C,\textsl{T})$ are ordered, and refer to the
$i$-th element of $\reconcepts (C,\textsl{T})$, $i=1\ldots \rbranchingfactor (C,\textsl{T})$, as $\rec _i(C,\textsl{T})$. Similarly, we
suppose that the abstract features in $\abstractfeatures (C,\textsl{T})$ are ordered, and refer to the
$i$-th element of $\abstractfeatures (C,\textsl{T})$, $i=1\ldots
\fbranchingfactor (C,\textsl{T})$, as $\af _i(C,\textsl{T})$. Together,
they constitute the directions of the weak alternating automaton to be associated with the satisfiability of $C$ $\wrt$ $\textsl{T}$.
\begin{definition}[branching tuple]\label{branchingfactor}
Let $C$ be an $\xdl$ concept and $\textsl{T}$ an $\xdl$ weakly cyclic TBox. The
branching tuple of $C$ $\wrt$ $\textsl{T}$ is given by the ordered $\branchingfactor (C,\textsl{T})$-tuple
$\branchingtuple (C,\textsl{T})=$\\
$(\rec _1(C,\textsl{T}),\ldots ,\rec _{\rbranchingfactor (C,\textsl{T})}(C,\textsl{T}),
                      \af _1(C,\textsl{T}),\ldots ,\af _{\fbranchingfactor (C,\textsl{T})}(C,\textsl{T}))$
                      of the $\rbranchingfactor (C,\textsl{T})$ relational existential concepts in $\reconcepts (C,\textsl{T})$ and
                      the $\fbranchingfactor (C,\textsl{T})$ abstract features in $\abstractfeatures (C,\textsl{T})$.
\end{definition}
Given an $\xdl$ concept $C$ and an $\xdl$ weakly cyclic TBox $\textsl{T}$, we will be
interested in $k$-ary $\Sigma$-trees (see Definition \ref{karymtree}), $t$,
verifying the following:
\begin{enumerate}
  \item $k=\branchingfactor (C,\textsl{T})$; and
  \item $\Sigma=2^{\primitiveconcepts (C,\textsl{T})}\times\Theta (\concretefeatures (C,\textsl{T}),\Delta _{\textsl{D}_x})$,
    where $\Theta (\concretefeatures (C,\textsl{T}),\Delta _{\textsl{D}_x})$ is the set of total functions
    $\theta :\concretefeatures (C,\textsl{T})\rightarrow\Delta _{\textsl{D}_x}$ associating with each concrete feature $g$ in
    $\concretefeatures (C,\textsl{T})$ a concrete value $\theta (g)$ from the spatial concrete domain  $\Delta _{\textsl{D}_x}$.
\end{enumerate}
Such a tree will be seen as representing a class of interpretations of the satisfiability of $C$ $\wrt$ $\textsl{T}$:
the label $(X,\theta )$ of a node $\alpha\in\{1,\ldots ,\branchingfactor (C,\textsl{T})\}^*$, with
$X\subseteq \primitiveconcepts (C,\textsl{T})$ and $\theta\in\Theta(\concretefeatures (C,\textsl{T}),\Delta _{\textsl{D}_x})$,
is to be interpreted as follows:
\begin{enumerate}
  \item $X$ records the information on the primitive concepts that are true at $\alpha$, in all interpretations of the class; and
  \item $\theta :\concretefeatures (C,\textsl{T})\rightarrow\Delta _{\textsl{D}_x}$ records the values, at the abstract object
    represented by node $\alpha$, of the concrete features $g_1,\ldots ,g_{\ncf (C,\textsl{T})}$
    in $\concretefeatures (C,\textsl{T})$.
\end{enumerate}
The crucial question is when we can say that an interpretation of the class is a model of $C$
$\wrt$ $\textsl{T}$.
To answer the question, we consider (weak) alternating automata on $k$-ary $\Sigma$-trees,
with $k=\branchingfactor (C,\textsl{T})$ and $\Sigma =2^{\primitiveconcepts (C,\textsl{T})}\times\Theta
(\concretefeatures (C,\textsl{T}),\Delta _{\textsl{D}_x})$.
We then show how to associate such an automaton with the satisfiability of an $\xdl$ concept $C$ $\wrt$ a weakly cyclic TBox $\textsl{T}$, in such a way
that the models of $C$ $\wrt$ $\textsl{T}$ coincide with the $k$-ary $\Sigma$-trees accepted by the automaton. The background on
alternating automata has been adapted from \cite{MullerSS92a}.
\section{Weak alternating automata and $\xdl$ with weakly cyclic Tboxes}
%
%
\begin{definition}[free distributive lattice]
Let $S$ be a set of generators. $\textsl{L}(S)$ denotes the free distributive lattice generated by $S$.
$\textsl{L}(S)$ can be thought of as the set of logical formulas built from variables taken from
$S$ using the disjunction and conjunction operators $\vee$ and $\wedge$ (but not the negation
operator $\neg$). In other words, $\textsl{L}(S)$ is the smallest set such that:
\begin{enumerate}
  \item for all $s\in S$, $s\in \textsl{L}(S)$; and
  \item if $e_1$ and $e_2$ belong to $\textsl{L}(S)$, then so do $e_1\wedge e_2$ and
    $e_1\vee e_2$.
\end{enumerate}
\end{definition}
Each element $e\in \textsl{L}(S)$ has, up to isomorphism, a unique representation in $\gdnf$
(Disjunctive Normal Form), $e=\bigvee _iC_i$ (each $C_i$ is a conjunction of generators
from $S$, and no $C_i$ subsumes $C_k$, with $k\not =i$). We suppose, without loss of generality, that
each element of $\textsl{L}(S)$ is written in such a form. If $e=\bigvee _i\bigwedge _js_{ij}$ is an
element of $\textsl{L}(S)$, the dual of $e$ is the element $\tilde{e}=\bigwedge _i\bigvee _js_{ij}$
obtained by interchanging $\vee$ and $\wedge$ ($\bigwedge _i\bigvee _js_{ij}$ is not necessarily in
$\gdnf$).
\begin{definition}[set representation]\label{setrepresentation}
Let $S$ be a set of generators, $\textsl{L}(S)$ the free distributive lattice generated by $S$, and
$e$ an element of $\textsl{L}(S)$. Write $e$ in DNF as $\bigvee _{i=1}^n\bigwedge _{j=1}^{n_i}s_{ij}$.
The set representation
of $e$, $\setrep (e)$, is the subset of $2^S$ defined as $\{S_1,\ldots ,S_n\}$, with
$S_i=\{s_{i1},\ldots ,s_{in_i}\}$.
\end{definition}
In the following, we denote
by $K$ a set of $k$ directions $d_1,\ldots ,d_k$; by $N_P$ a set of primitive concepts;
by $x$ an $\rcc8$-like $p$-ary spatial RA;
by $N_{cF}$ a finite set of concrete features referring to objects
in $\Delta _{\textsl{D}_x}$;
by $\alphabet$ the alphabet $2^{N_P}\times\Theta (N_{cF},\Delta
_{\textsl{D}_x})$,
$\Theta (N_{cF},\Delta _{\textsl{D}_x})$ being the set of total functions
$\theta :N_{cF}\rightarrow\Delta _{\textsl{D}_x}$, associating with each concrete feature $g$ a
concrete value $\theta (g)$ from the spatial concrete domain  $\Delta
_{\textsl{D}_x}$;
by $\lits (N_P)$ the set of literals derived from $N_P$ (viewed as a set of atomic
propositions):
$\lits (N_P)=N_P\cup\{\neg A:A\in N_P\}$;
by $c(2^{\lits (N_P)})$ the set of subsets of $\lits (N_P)$ which do
not contain a primitive concept and its negation: $c(2^{\lits
  (N_P)})=\{S\subset\lits (N_P):(\forall A\in N_P)(\{A,\neg
A\}\not\subseteq S)\}$;
by $\consts (x,K,N_{cF})$ the set of constraints of the form
$P(u_1,\ldots ,u_p)$ with $P$ being
an $x$ relation, $u_1,\ldots ,u_p$ $\ksgchains$ (i.e., $u_i$, $i\in\{1,\ldots ,p\}$, is of the form $g$ or $d_{i_1}\ldots d_{i_n}g$,
$n\geq 1$ and $n$ finite, the $d_{i_j}$'s being
directions in $K$, and $g$ a concrete feature).
\begin{definition}[B\"uchi alternating automaton]\label{buechialtaut}
Let $k\geq 1$ be an integer and $K=\{d_1,\ldots ,d_k\}$ a set of directions.
An alternating automaton on $k$-ary $\alphabet$-trees is a tuple
$\textsl{A}=(\textsl{L}(\lits (N_P)\cup\consts (x,K,N_{cF})\cup K\times Q),
           \alphabet ,$ $\delta ,q_0,\textsl{F})$,
where
$Q$ is a finite set of states;
$\alphabet$ is the input alphabet (labelling the nodes of the input trees);
$\delta :Q\rightarrow \textsl{L}(\lits (N_P)\cup\consts (x,K,N_{cF})\cup K\times Q)$ is the
transition function;
$q_0\in Q$ is the initial state; and
$\textsl{F}$ is the set of accepting states. $\textsl{A}$ is said to be a weak alternating automaton
if there exists a partial order $\geq$ on $Q$, so that the transition function $\delta$ has the property that, given two states
    $q,q'\in Q$, if $q'$ occurs in $\delta (q)$ then $q\geq q'$.
\end{definition}
Let $\textsl{A}$ be an alternating automaton on $k$-ary
$\alphabet$-trees, as defined in Definition \ref{buechialtaut}, and $t$ a $k$-ary
$\alphabet$-tree. Given two alphabets $\Sigma _1$ and $\Sigma _2$, we
denote by $\Sigma _1\Sigma _2$ the concatenation of $\Sigma _1$ and
$\Sigma _2$, consisting of all words $ab$, with $a\in\Sigma _1$
and $b\in\Sigma _2$. In a run $r(\textsl{A},t)$ of $\textsl{A}$ on $t$ (see below),
which can be seen as an unfolding of a branch of the computation tree
$T(\textsl{A},t)$ of $\textsl{A}$ on $t$, as defined in
\cite{MullerS87a,MullerSS92a,MullerS95a}, the nodes of level $n$ will represent one possibility
for choices of $\textsl{A}$ up to level $n$ in $t$.
For each $n\geq 0$, we define the set of
$n$-histories to be the set
$H_n=\{q_0\}(KQ)^n$ of all $2n+1$-length words
consisting of $q_0$ as the first letter, followed by a $2n$-length
word $d_{i_1}q_{i_1}\ldots d_{i_n}q_{i_n}$, with $d_{i_j}\in K$ and $q_{i_j}\in Q$, for all
$j=1\ldots n$. If $h\in H_n$ and
$g\in KQ$ then $hg$, the
concatenation of $h$ and $g$, belongs to $H_{n+1}$. More generally, if $h\in H_n$ and
$e\in \textsl{L}(KQ)$, the
concatenation $he$ of $h$ and $e$ will denote the element of $\textsl{L}(H_{n+1})$ obtained by
prefixing $h$ to each generator in
$KQ$ which occurs in $e$.
Additionally, given an $n$-history $h=q_0d_{i_1}q_{i_1}\ldots d_{i_n}q_{i_n}$, with $n\geq 0$, we denote:
\begin{enumerate}
  \item by $\last (h)$ the initial state $q_0$ if $h$ consists of the
    $0$-history $q_0$ ($n=0$),
and the state $q_{i_n}$ if $n\geq 1$;
  \item by $\kproj (h)$ (the $K$-projection of $h$) the empty word
    $\epsilon$ if $n=0$, and the $n$-length word $d_{i_1}\ldots d_{i_n}$ otherwise; and
  \item by $\qproj (h)$ (the $Q$-projection of $h$) the state $q_0$ if 
    $n=0$, and the $n+1$-length 
    word $q_0q_{i_1}\ldots q_{i_n}\in Q^{n+1}$ otherwise.
\end{enumerate}
The union of all $H_n$, with $n$ finite, will be referred to as the
set of finite histories of $\textsl{A}$, and denoted by $\hf$. We denote by $\alphabett$ the alphabet
$2^{\hf}\times c(2^{\lits (N_P)})\times 2^{\consts (x,K,N_{cF})}$,
by $\alphabettt$ the alphabet $2^Q\times c(2^{\lits (N_P)})\times
2^{\consts (x,K,N_{cF})}$, and, in general, by $\alphabtt$ the
alphabet $S\times c(2^{\lits (N_P)})\times
2^{\consts (x,K,N_{cF})}$.

A run of the alternating automaton $\textsl{A}$ on $t$ is now introduced.
\begin{definition}[Run]\label{definitionrun}
Let $\textsl{A}$ be an alternating automaton on $k$-ary
$\alphabet$-trees, as defined in Definition \ref{buechialtaut}, and $t$ a $k$-ary
$\alphabet$-tree. A run, $r(\textsl{A},t)$, of $\textsl{A}$ on $t$ is a
partial $k$-ary $\alphabett$-tree defined
inductively as follows. For all directions $d\in K$, and for all nodes $u\in K^*$ of
$r(\textsl{A},t)$, $u$ has at most one outgoing edge labelled with $d$, 
and leading to the $d$-successor $ud$ of $u$. The label
$(Y_{\epsilon},L_{\epsilon},X_{\epsilon})$ of the root belongs to $2^{H_0}\times c(2^{\lits (N_P)})\times 2^{\consts (x,K,N_{cF})}$
---in other words, $Y_{\epsilon}=\{q_0\}$. If $u$ is a
node of $r(\textsl{A},t)$ of level $n\geq 0$, with label
$(Y_u,L_u,X_u)$, then calculate
$e=\bigwedge _{h\in Y_u}\distribute (h,\delta (\last (h)))$,
where $\distribute$ is a function associating with each pair $(h_1,e_1)$
of $\hf\times \textsl{L}(\lits (N_P)\cup\consts (x,K,N_{cF})\cup
K\times Q)$ an element of $\textsl{L}(\lits (N_P)\cup\consts
(x,K,N_{cF})\cup\hf)$ defined inductively in the following way:\\
$
\distribute (h_1,e_1)=
  \left\{
                   \begin{array}{l}
                         e_1
                               \mbox{ if }e_1\in \lits (N_P)\cup\consts (x,K,N_{cF}),  \\
                         h_1dq
                               \mbox{ if }e_1=(d,q)\mbox{, with
                                 }(d,q)\in K\times Q,  \\
                         \distribute (h_1,e_2)\vee\distribute (h_1,e_3)
                               \mbox{ if }e_1=e_2\vee e_3,  \\
                         \distribute (h_1,e_2)\wedge\distribute (h_1,e_3)
                               \mbox{ if }e_1=e_2\wedge e_3  \\
                   \end{array}
  \right.
$\\
Write $e$ in $\dnf$ as
$e=\bigvee _{i=1}^r(L_i\wedge X_i\wedge Y_i)$, where the $L_i$'s are
conjunctions of literals from $\lits (N_P)$, the $X_i$'s are
conjunctions of constraints from $\consts
(x,K,N_{cF})$, and the $Y_i$'s are
conjunctions of $n+1$-histories.
Then there exists $i=1\ldots r$ such that
\begin{enumerate}
  \item $L_u=\{\ell\in\lits (N_P):\ell\mbox{ occurs in }L_i\}$;
  \item $X_u=\{x\in\consts (x,K,N_{cF}):x\mbox{ occurs in }X_i\}$;
  \item for all $d\in K$, such that the set
$Y=\{hdq\in H_{n+1}: (h\in H_n)\mbox{ and }(q\in Q)\mbox{ and } (hdq\mbox{ occurs in }Y_i)\}$
is nonempty, and only for those $d$, $u$ has a $d$-successor, $ud$, whose label
$(Y_{ud},X_{ud},L_{ud})$ is such that
$Y_{ud}=Y$; and
  \item the label $t(u)=(\textsl{P}_u,\theta _u)\in 2^{N_P}\times\Theta (N_{cF},\Delta
_{\textsl{D}_x})$ of the node
$u$ of the input tree $t$ verifies the following, where, given a
        node $v$ in $t$, the notation $\theta _v$ consists of the function $\theta _v:N_{cF}\rightarrow\Delta _{\textsl{D}_x}$ which is the second
        argument of $t(v)$:
    \begin{enumerate}
      \item[$\bullet$] for all $A\in N_P$: if $A\in L_u$ then $A\in \textsl{P}_u$; and
        if $\neg A\in L_u$ then $A\notin \textsl{P}_u$ (the elements $A$ of
        $N_P$ such that, neither $A$ nor $\neg A$ occur in $L_u$, may or may not
        occur in $\textsl{P}_u$);
      \item[$\bullet$] for all $P(d_{1_1}\ldots d_{1_{n_1}}g_1,\ldots ,d_{p_1}\ldots d_{p_{n_p}}g_p)$ appearing in $X_u$,\\
        $P(\theta _{ud_{1_1}\ldots d_{1_{n_1}}}(g_1),\ldots ,\theta _{ud_{p_1}\ldots d_{p_{n_p}}}(g_p))$ holds. In other words, the values of 
        the concrete features $g_i$, $i\in\{1,\ldots ,p\}$, at the $d_{i_1}\ldots
        d_{i_{n_i}}$-successors of $u$ in $t$ are
        related by the $x$ relation $P$.
    \end{enumerate}
\end{enumerate}
A partial $k$-ary $\alphabett$-tree $\sigma$ is a run of $\textsl{A}$ if there exists a $k$-ary
$\alphabet$-tree $t$ such that $\sigma$ is a run of $\textsl{A}$ on $t$.
\end{definition}
\begin{definition}[CSP of a run]\label{cspofarun}
Let $\textsl{A}$ be an alternating automaton on $k$-ary
$\alphabet$-trees, as defined in Definition \ref{buechialtaut}, and $\sigma$ a run of
$\textsl{A}$:
\begin{enumerate}
  \item for all nodes $v$ of $\sigma$, of label
$\sigma (v)=(Y_v,L_v,X_v)\in 2^{\hf}\times c(2^{\lits (N_P)})\times
2^{\consts (x,K,N_{cF})}$, the argument $X_v$ gives rise to the CSP of $\sigma$ at $v$,
    $\csp _v(\sigma )$, whose set of variables, $V_v(\sigma )$, and set of constraints,
    $C_v(\sigma )$, are defined as follows:
  \begin{enumerate}
    \item Initially, $V_v(\sigma )=\emptyset$ and $C_v(\sigma
      )=\emptyset$
    \item for all $\ksgchains$ $d_{i_1}\ldots d_{i_n}g$ appearing in
      $X_v$, create, and add to $V_v(\sigma )$, a variable $\langle vd_{i_1}\ldots d_{i_n},g\rangle$

    \item for all $P(d_{1_1}\ldots d_{1_{n_1}}g_1,\ldots ,d_{p_1}\ldots d_{p_{n_p}}g_p)$ in $X_v$, add the constraint\\
      $P(\langle vd_{1_1}\ldots d_{1_{n_1}},g_1\rangle,\ldots ,\langle vd_{p_1}\ldots d_{p_{n_p}},g_p\rangle)$ to $C_v(\sigma )$

  \end{enumerate}
  \item the CSP of $\sigma$, $\csp (\sigma )$, is the CSP whose set of variables,
    $\textsl{V}(\sigma )$, and set of constraints, $\textsl{C}(\sigma )$, are defined
    as $\textsl{V}(\sigma )=\displaystyle\bigcup _{v\mbox{ node of }\sigma}V_v(\sigma )$ and
    $\textsl{C}(\sigma )=\displaystyle\bigcup _{v\mbox{ node of }\sigma}C_v(\sigma )$.
\end{enumerate}
\end{definition}

An $n$-branch of a run $\sigma =r(\textsl{A},t)$ is a path of
length (number of edges) $n$ beginning at the root of
$\sigma$. A branch is an infinite path. If $u$ is the terminal node of an
$n$-branch $\beta$, then the argument $Y_u$ of the label $(Y_u,L_u,X_u)$ of
$u$ is a set of $n$-histories. Following \cite{MullerSS92a}, we say that
each $n$-history in $Y_u$ lies along $\beta$. An $n$-history $h$ lies
along $\sigma$ if there exists an $n$-branch $\beta$ of $\sigma$ such
that $h$ lies along $\beta$.
An (infinite) history is a sequence
$h=q_0d_{i_1}q_{i_1}\ldots d_{i_n}q_{i_n}\ldots\in\{q_0\}(KQ)^\omega$.
Given such a history,
$h=q_0d_{i_1}q_{i_1}\ldots d_{i_n}q_{i_n}\ldots\in\{q_0\}(KQ)^\omega$:
\begin{enumerate}
  \item $h$ lies along a branch $\beta$ if, for every $n\geq 1$, the prefix of $h$
    consisting of the $n$-history
    $q_0d_{i_1}q_{i_1}\ldots d_{i_n}q_{i_n}$ lies along the
    $n$-branch $\beta _n$ consisting of the first $n$ edges of
    $\beta$;
  \item $h$ lies along $\sigma$ if there exists a branch $\beta$ of
    $\sigma$ such that $h$ lies along $\beta$;
  \item $\qproj (h)$ (the $Q$-projection of $h$) is the infinite word
    $q_0q_{i_1}\ldots q_{i_n}\ldots \in Q^\omega$ such that, for all $n\geq 1$, the
    $n+1$-length prefix $q_0q_{i_1}\ldots q_{i_n}$ is the $Q$-projection of
    $h_n$, the $n$-history which is the $2n+1$-prefix of $h$.
  \item we denote by $\infinity (h)$ the set of
states appearing infinitely often in $\qproj (h)$
\end{enumerate}
The acceptance condition is now defined as follows. A history $h$ is accepting if
$\infinity (h)\cap \textsl{F}\not =\emptyset$. A branch
$\beta$ of $r(\textsl{A},t)$ is accepting if every history lying along
$\beta$ is accepting.

The condition for a run $\sigma$ to be accepting splits into two subconditions. The
first subcondition is the standard one, and is related to (the
histories lying along) the
branches of $\sigma$, all of which should be accepting. The second subcondition
is new: the CSP of $\sigma$, $\csp (\sigma )$,
should be consistent. $\textsl{A}$ accepts a $k$-ary $\alphabet$-tree $t$ if there exists an accepting
run of $\textsl{A}$ on $t$. The language $\textsl{L}(\textsl{A})$ accepted by $\textsl{A}$ is
the set of all $k$-ary $\alphabet$-trees accepted by $\textsl{A}$.
\section{Associating a weak alternating automaton with the satisfiability of a concept $\wrt$ a weakly cyclic TBox}
Summarising the previous steps, especially the work of the procedure of Figure \ref{dnf2procedure}, we get the following corollary.
\begin{corollary}
Let $x$ be a spatial RA of arity $p$,
$C$ an $\xdl$ concept,
$\textsl{T}$ an $\xdl$ weakly cyclic TBox, $\tpc$ the TBox $T$ augmented 
with $C$, and $B_i$ the initial defined concept of $\tpc$.
$C$
is satisfiable $\wrt$ $\textsl{T}$ $\iff$ the language $\textsl{L}(\textsl{A}_{C,\textsl{T}})$
accepted by weak alternating automaton
$\textsl{A}_{C,\textsl{T}}=(\textsl{L}(\lits (N_P)\cup\consts (x,K,N_{cF})\cup K\times Q),
           \alphabet ,\delta ,q_0,\textsl{F})$ on $k$-ary $\alphabet$-trees is nonempty. The parameters of the automaton are as
follows:
\begin{enumerate}
  \item $N_P=\primitiveconcepts (C,\textsl{T})$,
    $N_{cF}=\concretefeatures (C,\textsl{T})$,
    $Q=\definedconcepts (C,\textsl{T})$, $q_0=B_i$
  \item $K$ is the set of relational existential concepts and abstract features appearing as arguments in the branching tuple of $C$ $\wrt$ $\textsl{T}$:
    $K=\{d_1,\ldots ,d_n:(d_1,\ldots ,d_n)=\branchingtuple (C,\textsl{T})\}$ (Definition
    \ref{branchingfactor})
  \item $\delta (B)$ is obtained from the axiom $B\doteq E$ in $(\tpc
    )^*$ defining $B$, as follows. $E$ is of the form $\{S_1,\ldots
    ,S_n\}$, with $S=S_{prop}\cup S_{csp}\cup S_{\exists}$, for all
    $S\in\{S_1,\ldots ,S_n\}$.
    \begin{enumerate}
      \item  We transform $E$ into $E'=\{\mu (S_1),\ldots ,\mu (S_n)\}$, with $\mu (S)$, $S\in\{S_1,\ldots ,S_n\}$, computed as follows:
        \begin{enumerate}
      \item Let
               $Set_{\exists}=\{(\exists R.D,B_1)\wedge\cdots\wedge (\exists R.D,B_{\ell}):\mbox{ $R$ general role and }\exists R.D\in S_{\exists}\mbox{ and }
D=B_1\sqcap\cdots\sqcap B_{\ell}\}\cup
                                      \{(f,B_1)\wedge\cdots\wedge (f,B_{\ell}) :\mbox{ $f$ abstract feature and }
\exists f.D\in S_{\exists}\mbox{ and }
D=B_1\sqcap\cdots\sqcap B_{\ell}\}$.
      \item Let
$
Set_{csp}=
                         \{P(u_1,\ldots ,u_p):u_1,\ldots u_p\in K^*N_{cF}\mbox{ and }\exists (u_1)\ldots (u_p).P\in S_{csp}\}
$
      \item Let $\mu (S)=S_{prop}\cup Set_{csp}\cup Set_{\exists}$.
        \end{enumerate}
      \item We now have $\delta
    (B)=\displaystyle\bigvee _{S\in E}\bigwedge _{X\in\mu (S)}X$.
    \end{enumerate}
  \item The set $F$ of accepting states is the set of defined concepts in $\definedconcepts (C,\textsl{T})$ that are not evenuality defined concepts
  \item Finally, the partial order $\geq$ on the states in $Q$ is as computed by the procedure of of Figure \ref{dnf2procedure}. \cqfd
\end{enumerate}
\end{corollary}

\section{Conclusion and future work}\label{conclusion}
We have investigated a
spatio-temporalisation $\xdl$ of the $\alcd$ family of description logics with a concrete
domain \cite{BaaderH91a}, obtained by temporalising the roles, so that
they consist of $m+n$ immediate-successor (accessibility) relations,
the first $m$ being general, the
other $n$ functional; and spatialising the concrete domain,
which is generated by an $\rcc8$-like qualitative spatial language
\cite{RandellCC92a,Egenhofer91a}.

We have shown the important result that satisfiability of an $\xdl$ concept with respect to a weakly cyclic TBox
can be reduced to the emptiness problem of a B\"uchi weak alternating automaton augmented with
spatial constraints.
%
%

In another
work, complementary to this one, also submitted to this conference, we thoroughly investigate B\"uchi
automata augmented with spatial constraints, and provide, in particular, a translation of an alternating
into a nondeterministic, and a nondeterministic doubly depth-first polynomial space algorithm for the emptiness problem of the latter.
Together, the two works provide an effective
solution to the satisfiability problem of an $\xdl$ concept with respect to a weakly cyclic TBox.

A future work worth mentioning is whether
one can keep the same spatio-temporalisation and define a form of TBox cyclicity
stronger than the one considered in this work, and expressive enough to subsume the semantics of the well-known mu-calculus.

\newpage\noindent
\bibliographystyle{aaai}
\bibliography{biblio-c-maj}

\end{document}